\documentclass[10pt,twocolumn,letterpaper]{article}

\usepackage{wacv}
\usepackage{times}
\usepackage{epsfig}
\usepackage{graphicx}
\usepackage{amsmath}
\usepackage{amssymb}
\usepackage{booktabs}
\usepackage{dcolumn}
\usepackage[percent]{overpic}

\wacvfinalcopy % *** Uncomment this line for the final submission

\def\wacvPaperID{627} % *** Enter the wacv Paper ID here

\ifwacvfinal\fi
\begin{document}

%%%%%%%%% TITLE
\title{Gyroscope-Aided Motion Deblurring with Deep Networks}

% Authors at the same institution
%\author{First Author \hspace{2cm} Second Author \\
%Institution1\\
%{\tt\small firstauthor@i1.org}
%}
% Authors at different institutions
%\author{First Author \\
%Institution1\\
%{\tt\small firstauthor@i1.org}
%\and
%Second Author \\
%Institution2\\
%{\tt\small secondauthor@i2.org}
%}

\author{$^1$Janne Mustaniemi \hspace{0.5cm} $^2$Juho Kannala \hspace{0.5cm} $^2$Simo S\"arkk\"a \hspace{0.5cm} $^3$Jiri Matas \hspace{0.5cm} $^1$Janne Heikkil\"a\\ \small $^1$Center for Machine Vision and Signal Analysis, University of Oulu, Finland \hspace{0.5cm} \small $^2$Aalto University, Finland\\ \small $^3$Center for Machine Perception, Faculty of Electrical Engineering, Czech Technical University in Prague, Czech Republic\\ {\tt\small janne.mustaniemi@oulu.fi}
}

%\author{Janne Mustaniemi\\
%University of Oulu\\Finland\\
%{\tt\small janne.mustaniemi@oulu.fi}
%\and
%Juho Kannala\\ and Simo S\"arkk\"a\\
%Aalto University\\Finland
%\and
%Jiri Matas\\
%Czech Technical\\University
%in Prague\\Czech Republic
%\and
%Janne Heikkil\"a\\
%University of Oulu\\Finland
%}

\maketitle
\ifwacvfinal\thispagestyle{empty}\fi

%========================================================================
\begin{abstract}
%Motion deblurring methods have improved considerably over recent years, largely due to advances in deep learning. Promising results have also been obtained with non-learning based approaches that utilize inertial measurements. 
We propose a deblurring method that incorporates gyroscope measurements into a convolutional neural network (CNN). With the help of such measurements, it can handle extremely strong and spatially-variant motion blur. At the same time, the image data is used to overcome the limitations of gyro-based blur estimation. To train our network, we also introduce a novel way of generating realistic training data using the gyroscope. The evaluation shows a clear improvement in visual quality over the state-of-the-art while achieving real-time performance. Furthermore, the method is shown to improve the performance of existing feature detectors and descriptors against the motion blur.
\end{abstract}

%========================================================================
\section{Introduction}
Motion blur is often unavoidable when capturing images with a fast moving camera. It not only degrades the visual quality but it also has a negative impact on applications such as visual odometry, augmented reality (AR) and simultaneous localization and mapping (SLAM). Even though the blind deblurring methods have improved significantly over the years, they generally struggle with strong and spatially-variant motion blur. We intend to overcome these limitations by utilizing inertial measurements.

\begin{figure}
\centering
\includegraphics[width=8.2cm]{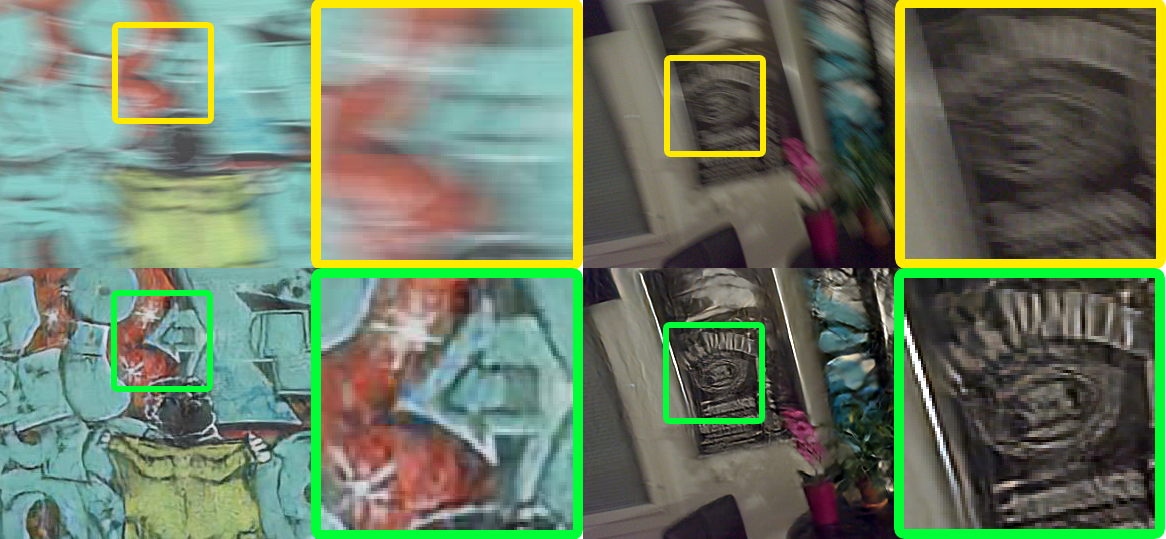}
\caption{Heavily blurred images captured with a fast moving camera (top). Images deblurred
by DeepGyro CNN (bottom). }
\label{fig:teaser}
\end{figure}

Blind deconvolution methods aim to recover the sharp image without any additional information about the motion blur. This is an ill-posed problem since the blurred image only provides a partial constraint on the solution. Promising results have been obtained with recent deep learning based approaches \cite{DeblurGAN,nah2017deep}. These methods are especially good at generating perceptually convincing images while avoiding deblurring artifacts. To simplify the problem, the existing methods typically assume a spatially-invariant blur, which may not hold in practice. An example of such case is shown in Figure \ref{fig:teaser}. 

Mobile devices are often equipped with an inertial measurement unit (IMU), which provides information about the motion blur. Accelerometers and gyroscopes have been successfully used in motion deblurring \cite{joshi2010image,vsindelavr2013image,hee2014gyro,hu2016image,zhang2016combining,mustaniemi2018fast}. Most of these methods focus on the removal of the camera shake blur. An application such as SLAM may involve a fast moving camera, which generally results in much stronger motion blur. The existing methods are also not capable of running in real-time, apart from \cite{mustaniemi2018fast}. What further complicates the problem is that inertial-based blur estimates may be inaccurate. This can be due to noisy IMU readings, temporal misalignment between the camera and IMU, unknown scene depth or translation. These limitations should be considered in order to avoid deblurring artifacts.

We propose a deblurring method that incorporates gyroscope measurements into a convolutional neural network (CNN). It can handle extremely strong and spatially-variant motion blur as illustrated in Figure \ref{fig:teaser}. When computing the gyro-based blur estimates, we take into account that mobile devices are usually equipped with a rolling shutter camera. The method naturally overcomes the limitations of gyro-based blur estimation by utilizing image data. We also introduce a novel data generation scheme, which is an essential component needed to train our network. The evaluation on real-world images shows a clear improvement in visual quality over the state-of-the-art while achieving real-time performance. The method will also improve the robustness of existing feature detectors and descriptors against motion blur as indicated by the higher repeatability and better matching performance. 

% NOTES
% - Motion blur caused by camera shake vs. intentionally moved camera
% - Decoupled blur estimation / deblurring
% - It is non-trivial to use inertial measurements in a CNN

%=========================================================================
\section{Related work}

%Paragraph 1 - Blind deblurring. Traditional methods \cite{pan2016blind,yan2017image}, Learning-based \cite{DeblurGAN,nimisha2017blur}.
%\\ ...

Despite being a classical image processing problem, deblurring continues to be an active research area with plenty of recent progress. For example, regarding blind single-image deblurring, the recent papers utilizing so called dark and bright channel priors have shown promising results \cite{pan2016blind,yan2017image}. Nevertheless, these approaches typically assume uniform and spatially invariant blur, which is often not the case in practice. For example, if there is rotation around optical axis, the blur kernel is clearly spatially variant. 

Recently, there have emerged also several deep learning based blind deblurring methods. For example, the concept of generative adversarial networks has been utilized for learning deep neural networks that perform deblurring \cite{DeblurGAN,nimisha2017blur,nah2017deep}. In particular, inspired by \emph{pix2pix} \cite{pix2pix2017}, \emph{DeblurGAN} \cite{DeblurGAN} trains a conditional GAN for deblurring using pairs of corresponding blurred and sharp images. However, as the blind deblurring problem is severely ill-posed, the results are often not good or satisfactory. In fact, we use \emph{DeblurGAN} as one of the baseline methods and Figures \ref{fig:synthetic} and \ref{fig:natural} illustrate that its results are clearly inferior to ours.

Besides methods that directly perform blind deblurring, there are also approaches that first estimate a spatially-variant motion field and blur kernels from a single image using deep networks, and thereafter perform non-blind deconvolution \cite{sun2015learning,gong2017motion}. Further, deep nets have been trained to remove deblurring artifacts that non-blind deconvolution typically creates, either directly predicting the sharp output image \cite{son2017fast} or the residual image between the deconvolution result and the desired sharp output \cite{wang2018training}.   

%Paragraph 2 - Blur estimation and non-blind deconvolution. Spatially-variant blur estimation followed by non-blind deconvolution \cite{sun2015learning,gong2017motion},
%Non-blind deconvolution with deep learning \cite{son2017fast,wang2018training}.
%\\ ...

In addition to single-image deblurring methods, there are also methods that utilize additional information, such as multiple frames from a video \cite{chen2018reblur,su2017deep}, bursts of rapidly captured photographs \cite{aittala2018burst}, pairs of blurred and noisy images captured with different exposure settings \cite{yuan2007image}, or high- and low-resolution image pairs \cite{tai2010correction}. While some of the aforementioned methods provide promising results, they belong to a different domain than our single-image deblurring approach. Moreover, multiple images are not always available or easy to capture as dynamic objects and events may disappear from the scene.% Also, if there is a short time budget for exposure in low-light conditions, it may be better to capture a single long-exposure frame instead of several short-exposure frames in order to avoid sensor noise and delays/overheads due to shutter speed and storage of multiple images.

%Paragraph 3 - Methods that utilize additional information. Video deblurring \cite{chen2018reblur,su2017deep}, Burst deblurring \cite{aittala2018burst}, Blurred and noisy image pairs \cite{yuan2007image}, High and low-resolution image pairs \cite{tai2010correction}.
%\\ ...

Our work deals with inertial-aided single-image deblurring. That is, we learn a deep neural net for deblurring a single RGB image so that the input to the net is the blurred image and a spatially varying motion field estimated based on gyroscope measurements recorded during the exposure of the image. This problem setting is highly relevant in practice since rotation is usually the main source of blur due to hand shake \cite{hee2014gyro} and most smartphones are equipped with gyroscopes. %that measure the angular rate of rotation with a high frequency. 
There have been relatively many papers that utilize inertial sensors (gyroscopes and/or accelerometers) for image deblurring \cite{joshi2010image,hu2016image,hee2014gyro,zhang2016combining,mustaniemi2018fast,vsindelavr2013image}. Most of them focus on estimating and characterizing the blur kernels based on the inertial sensor data \cite{joshi2010image,hu2016image,hee2014gyro,mustaniemi2018fast,vsindelavr2013image} and then apply non-blind deconvolution. Nevertheless, due to the limitations of consumer grade inertial sensors in smartphones, the motion estimates can never be perfect and, in practice, there may also be dynamic objects in the scene and their apparent motion is not explained by device motion. Thus, it seems plausible to combine inertial measurements and image based information for deblurring \cite{zhang2016combining} and our work does that by utilizing deep CNNs. To the best of our knowledge, our method is the first one that combines gyroscope measurements and learnt neural network based image priors for deblurring. This approach has significant benefits as our results show a clear improvement in visual quality over the previous state-of-the-art %\cite{mustaniemi2018fast} 
 while achieving real-time performance.

%Paragraph 4 - Inertial-based deblurring \cite{joshi2010image,hu2016image,hee2014gyro,zhang2016combining,mustaniemi2018fast,vsindelavr2013image}
%\\ ...

%
%-------------------------------------------------------------------------
\section{Blur estimation} \label{sec:blur_estimation}
Motion blur is caused by the relative motion of the camera and scene during the exposure of the image. This work focuses on static scenes, meaning the motion blur is only due to the rotation and translation of the camera. %Figure \ref{fig:overview} shows the overview of the proposed method.
The initial estimate for the motion blur is obtained with the gyroscope. A key challenge is to represent this information in a format useful for the deep network. This process will be covered in the next section. As a result, we get a spatially-variant blur field, which is provided for the deblurring network as an additional input.

%-------------------------------------------------------------------------
%\subsection{Estimation of rotation from gyroscope measurements}
\subsection{Rotation from gyroscope measurements}
In prior work \cite{vsindelavr2013image,hee2014gyro}, it has been shown that motion blur is typically caused by the rotation of the camera. Similar to these works, we compute the rotations by integrating gyroscope readings. More specifically, we numerically integrate the quaternion
differential equation (e.g. \cite{Titterton+Weston:2004})
\begin{equation}
  \frac{d\mathbf{q}(t)}{dt} = \frac{1}{2} \, \mathbf{q}(t) \, \odot \omega(t), \quad \mathbf{q}(t_1) = \mathbf{1},
\label{eq:qdiff}
\end{equation}
where $\omega(t)$ is the 3-dimensional gyroscope measurement and $\odot$ denotes the quaternion product. The initial condition is given at the starting time of exposure $t_1$ and the solution is computed at the end  time of exposure $t_2$. The rotation matrix $\mathbf{R}(t_2)$ is then formed as the direction cosine matrix corresponding to the quaternion $\mathbf{q}(t_2)$ (see, e.g., \cite{Titterton+Weston:2004}).

In theory, the translation could also be recovered using an accelerometer \cite{zhang2016combining,hu2016image,joshi2010image,Solin:2018}. However, this requires knowledge of the initial velocity of the camera, or alternatively, known stationary points or reference points which can be used to aid zero-velocity updates or position updates in a Kalman filter \cite{Solin:2018}. However, these are not assumed to be available here. Furthermore, the motion blur caused by translation will also depend on the scene depth, which is difficult to estimate from a single image. We take these limitations into account when generating training data.

\subsection{Blur field computation}
If the camera is moving during the image exposure, the 3D scene points will be projected to multiple points on the image plane. This will appear as motion blur. To estimate the blur, we need to consider the relative motion of the camera during the exposure. Let $\mathbf{R}(t)$ and $\mathbf{t}(t)$ denote the rotation and translation of the camera. Assuming that the scene has a constant depth $d$, the motion can be modeled using a planar homography \cite{hartley2003multiple}
\begin{equation}
\mathbf{H}(t) = \mathbf{K} [ \mathbf{R}(t) - \frac{\mathbf{t}(t) \mathbf{n}^{\top}}{d} ] \mathbf{K}^{-1},
\label{eq:homography}
\end{equation}
where $\mathbf{K}$ is the intrinsic camera matrix obtained via calibration. The normal vector of the scene is denoted by $\textbf{n}$. If the translation is zero (or if the scene is far away), the previous equation simplifies to
\begin{equation}
\mathbf{H}(t) = \mathbf{K} \mathbf{R}(t) \mathbf{K}^{-1}
\label{eq:homography_simple}
\end{equation}
Let $\mathbf{x}=(x,y,1)^\top$ be the projection of the 3D point at the beginning of the exposure. The rest of the projections can be computed by $\mathbf{x}' = \mathbf{H}(t) \mathbf{x}$.

\begin{figure}
\centering
\begin{overpic}[width=0.45\textwidth]{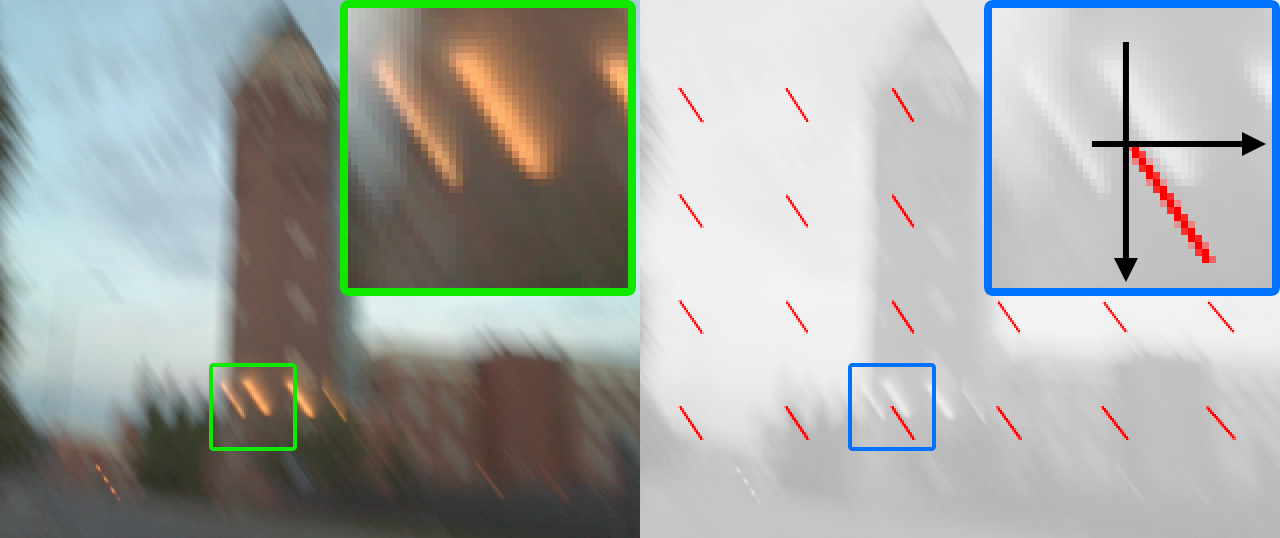}
\put (96,33.0) {$u$}
\put (83.0,20.5) {$v$}
\put (77.3,33.0) {$(0,0)$}
\end{overpic}
\caption{Comparison of the real blur and gyro-based blur kernel estimates $(u,v)$, in red. The local real blur is best visible at light-streaks.}
\label{fig:blur_estimation}
\end{figure}

If the exposure time is relatively short (e.g. when capturing a video), the motion blur can be assumed to be linear and homogeneous. This type of blur can be described with a 2-dimensional blur vector $(u,v)$, where $u$ and $v$ represent the horizontal and vertical components of the blur, respectively. See the visualization in Figure \ref{fig:blur_estimation}. %Similar notation was used in \cite{sun2015learning} and \cite{gong2017motion}. 
Note that all blur vectors with equal lengths and opposite directions, such as $(u,v)$ and $(-u,-v)$ will correspond to the same blur. Therefore, we choose to constrain the horizontal component $u$ to be positive. We compute the blur vectors for every pixel, which gives us the blur maps $\mathbf{U}$ and $\mathbf{V}$ in horizontal and vertical directions. Together these are referred to as blur field $\mathcal{B} = (\mathbf{U},\mathbf{V})$.

\subsection{Rolling shutter effect}
Mobile devices are typically equipped with a rolling shutter camera. This means, each row of pixels will be captured at slightly different time. The formula \ref{eq:homography_simple} cannot be used directly since the mapping of the point $\mathbf{x}$ depends on its y-coordinate. Let $t_r$ denote the camera readout time, that is the time difference between the exposure of the first and last row of pixels. Then, the exposure of the $y$:th row starts at
\begin{equation}
t_1(y) = t_f + t_r \frac{y}{N},
\label{eq:rolling_shutter}
\end{equation}
where $t_f$ is the frame timestamp and $N$ is the number of rows. The end of the exposure is defined as $t_2 = t_1 + t_e$, where $t_e$ is the exposure time. The mapping of the point then becomes
\begin{equation}
\mathbf{x}' = \mathbf{K}\mathbf{R}(t_2)\mathbf{R}^{\top}(t_1)\mathbf{K}^{-1}\mathbf{x}.
\label{eq:homography_rolling}
\end{equation}
Note that the frame timestamp $t_f$, readout time $t_r$ and exposure time $t_e$ can be typically obtained via the API of the mobile device.

\begin{figure*}[!htbp]
\centering
\includegraphics[width=17.0cm]{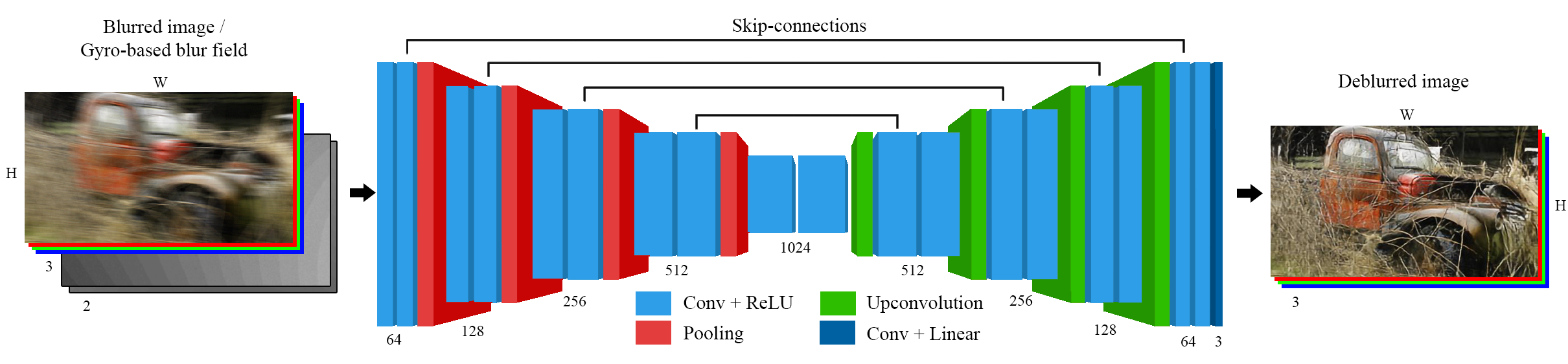}
\caption{The DeepGyro CNN architecture. All convolutional layers use a 3x3 window (except the last one, which is 1x1). The number of channels is shown below the boxes. Downsampling is 2x2 max pooling with stride 2. Upconvolutional layers consist of upsampling and 2x2 convolution that halves the number of feature channels.}
\label{fig:overview}
\end{figure*}

%-------------------------------------------------------------------------
\section{Deblurring}
Deblurring is based on a fully-convolutional neural network. It aims to produce a sharp image given a blurred image and gyro-based blur field. The architecture of the network is described in the next section. To train the network, we propose a data generation scheme that utilizes gyroscope readings. This topic is covered in Section \ref{sec:training}.

\subsection{Network architecture}
% Paragraph 1 - Related work and reason behind the chosen architecture
%------------------------------------------------------------------------
%
%
% Paragraph 2 - Inputs, outputs and details
%------------------------------------------------------------------------
% Inputs: Blurred RGB image (M x N x 3) + Gyro-based blur field (M x N x 2)
% Output: Deblurred RGB image (M x N x 3)
% 
The architecture of the network is shown in Figure \ref{fig:overview}. The network is similar to U-Net \cite{ronneberger2015u}, which was originally used for image segmentation. This type of encoder-decoder network has proven to be useful in various image-to-image translation problems \cite{pix2pix2017}. The input of our network consists of a blurred RGB image and a gyro-based blur field. They pass through a series of convolutional and downsampling layers, until the lowest resolution is reached. After the bottleneck, this process is reversed. A low-resolution image is expanded back into a full resolution image with help of upsampling layers. Skip connections are used to allow information sharing between the encoder and decoder. Given two layers with equal size, the feature maps from the encoder are concatenated with those of the decoder. The input images can be of arbitrary size since the network is fully-convolutional.

\subsection{Data generation} \label{sec:training}

\begin{figure*}[!h]
\centering
\begin{overpic}[width=0.95\textwidth]{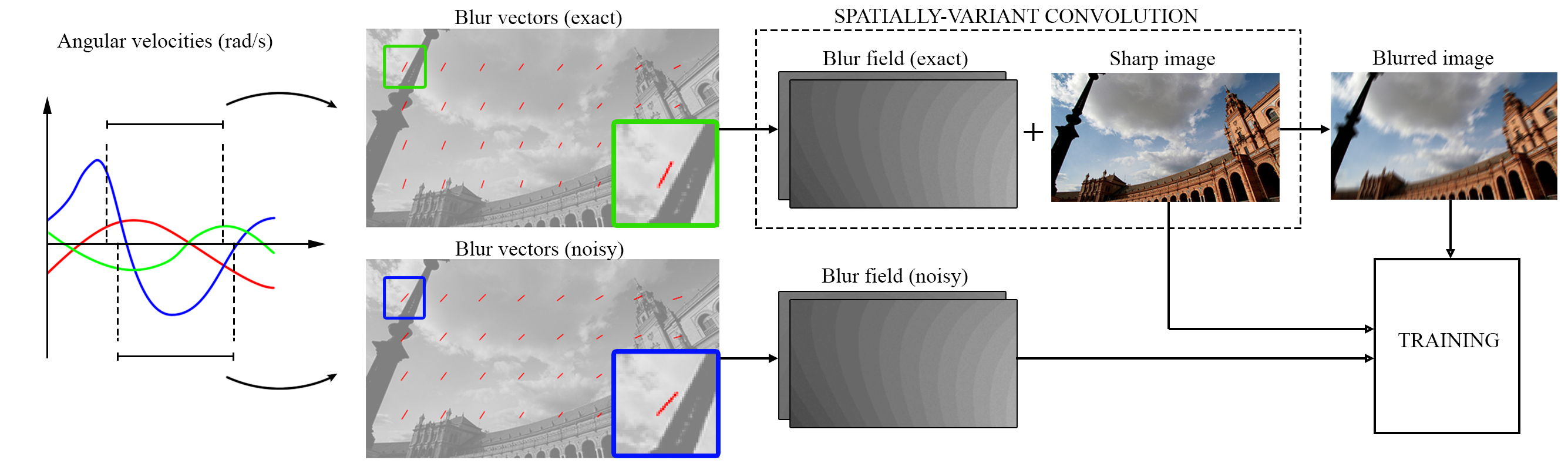}
\put (5.8,23.5) {$t_f$}
\put (4.5,5) {$t_f+t_d$}
\end{overpic}
\caption{DeepGyro training. For each image, two slightly different blur fields are generated, one that is applied to the sharp image (exact blur, top) and a noisy blur, modelling the IMU derived field (noisy, bottom).}
%\caption{The overview of the data generation scheme. For each image, we generate two slightly different blur fields. See the green and blue patches. We use the first blur field (exact) to generate the blurred image. The second blur field (non-idea) is given for the network as input.}
\label{fig:data_generation}
\end{figure*}

% Paragraph 1 - Goal and motivation
%------------------------------------------------------------------------
% - To train the network, we need: blurred images, spatially-variant blur fields and sharp images. 
% - Motivation: Capturing real-world data is difficult, Synthetic data needs to be realistic. Current approaches have limitations.
% - The training set should consist of realistically blurred images.
% - Piece-wise smooth
% - We generate realistic blur fields using the gyroscope readings.

%To train the network, we need a set of blurred and sharp images along with gyro-based blur fields. There is no easy way to capture corresponding pairs of blurred and sharp images. Noroozi et al. \cite{noroozi2017motion} and Nah et al. \cite{nah2017deep} generate such image pairs with a high frame-rate camera. After capturing a sequence of frames, they use the central frame as the sharp image and the average of the frames as the blurred image. To generate realistically blurred images, they avoid large motion between subsequent frames. In our case, the camera might be moving very fast. Even if the camera was recording at 240 fps, the pixel displacements could be much larger than 1 pixel. Another problem is that gyro-based blur fields are also required to train our network.

To train the network, we need a set of blurred and sharp images along with gyro-based blur fields. There is no easy way to capture such real-world data. As mentioned, the motion blur is mainly caused by the rotation of the camera. We utilized gyroscope readings to generate realistic blur fields and blurred images. Specifically, we use the sequences \textit{room1} - \textit{room6} from an existing visual-inertial dataset \cite{schubert2018vidataset}. These sequences consist of various types of camera motion, which results to a diverse set of blur fields with varying levels of spatially-variant motion blur. We also utilize images from the Flickr image collection \cite{huiskes2010new} to cover a wide range of different scene types. With the proposed data generation scheme, it is easy to generate practically unlimited amount of realistic training data. The data generation tool will be made publicly available upon the publication of the paper.
% (In our case, the exposure time is 30 ms or less. During this time, a high frame-rate camera (recording at 240 fps), could only capture a few frames. This would not be enough -> very large displacements.)

% - General description
% - Good things: Generated motion blur is piece-wise smooth and realistic. Allows us to generate gyro-based blur fields, which are needed to train our network. We can naturally introduce errors to gyro-based blur fields to simulate limitations of inertial-based blur estimation. Diverse training set: different amounts of spatially-variant blur, not forgetting sharp images. Training set is easy to scale since we can simply get more images from the Internet. Could be easily extended to generate PSFs of arbitrary shape.

%As mentioned previously, the motion blur is mainly caused by the rotation of the camera. We utilized gyroscope readings to generate realistic blur fields and blurred images. Specifically, we use the sequences \textit{room1} - \textit{room6} from an existing visual-inertial dataset \cite{schubert2018vidataset}. These sequences consist of various types of camera motion, which results to a diverse set of blur fields with varying amounts of spatially-variant motion blur. We also utilize images from the Flickr image collection \cite{huiskes2010new} to cover a wide range of different scene types. With the proposed data generation scheme, it is easy to generate practically unlimited amount of realistic training data. The data generation tool will be made publicly available upon the publication of the paper.

The overview of the data generation scheme is shown in Figure \ref{fig:data_generation}. We compute two different blur fields, which we refer as the "exact" and "noisy" blur fields. The exact blur field is used for generating the blurred image. We perform a spatially-variant convolution given a sharp image and blur kernels for every pixel. The noisy blur field, which is slightly different, is provided for the deblurring network as an additional input.

To generate a blur field, we use the approach described in Section \ref{sec:blur_estimation}. The start of the exposure $t_f$ is selected randomly, which means every blur field is likely to be somewhat different. We set the exposure time $t_e = 30$ milliseconds. The readout time $t_r$ is chosen randomly from the range [0,30] milliseconds. The zero value corresponds to a global shutter camera. To increase the overall level of motion blur, the angular velocities were first multiplied by 2. However, the maximum blur was limited to 100 pixels. 

To simulate temporal misalignment between the camera and gyroscope, we add a small delay $t_d$ to the start of the exposure when computing the noisy blur field. The delay is sampled from normal distribution with $\mu = 0$ and $\sigma=0.01$ milliseconds. The translation will also affect the motion blur if the scene is close to the camera. In such case, the blur extents observed by the gyroscope will be somewhat inaccurate. To this end, we multiply the gyroscope readings with a small number $k \sim N(0,0.2)$ before computing the noisy blur field. This will mainly affect the blur extents, rather than the direction of the blur.

\subsection{Training}
DeepGyro was trained on 100k images with resolution of 270 $\times$ 480 pixels. We used the Adam \cite{kingma2015j} as the solver. At the beginning, the learning rate was set to 0.00005. After every 10-th epoch, the learning rate was halved. The network was trained for 40 epochs. For comparison, we also trained a blind deblurring network, which we refer to as DeepBlind. In contrast to DeepGyro, it does not take the blur field as input. The network and training details are otherwise identical.

%There is a total of13.5 minutes of gyroscope readings, sampled at 200 Hz

%A sharp image taken from the Flickr image collection is then blurred using the computed blur field. We use total of 100k images for training. After blurring, the images were resized and cropped to size of 270 $\times$ 480 pixels.

%-------------------------------------------------------------------------
\section{Experiments}
Deblurring performance is evaluated on both synthetically and naturally blurred images. We compare the proposed approaches against DeblurGAN \cite{DeblurGAN} and Mustaniemi \etal \cite{mustaniemi2018fast}. DeblurGAN is a blind deblurring method based on the conditional generative adversarial networks. Similar to our DeepBlind approach, it only takes the blurred image as input. The gyro-based deblurring method \cite{mustaniemi2018fast}, referred to as FastGyro, is the closest competitor to our DeepGyro approach. We use a slightly modified version of the original implementation. The blur kernels are estimated for each pixel instead of image patches. This minimizes the artifacts near the edges of the patches. %In the last section, we use the proposed methods for feature detection and matching.

\subsection{Synthetic blur}
For the quantitative comparison, we add synthetic motion blur and 30 dB Gaussian noise to sharp images \cite{mikolajczyk2005performance}. The evaluation metrics include peak-signal-to-noise ratio (PSNR) and structural similarity (SSIM). For fairness, the motion blur is spatially-invariant since DeblurGAN \cite{DeblurGAN} is not designed to handle spatially-variant blur. Note that we also need to generate noisy blur fields for the non-blind methods because the gyroscope readings do not really exist.

\begin{figure*}[!h]
\centering
\begin{overpic}[width=1.00\textwidth]{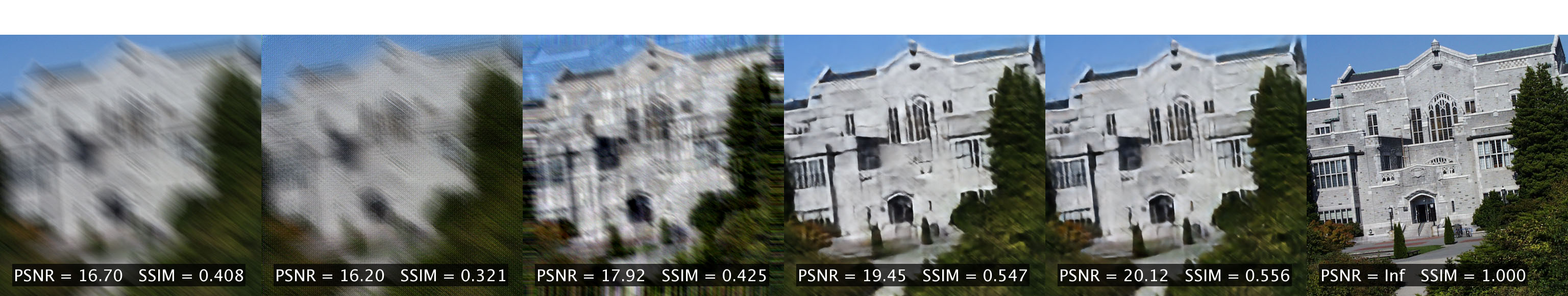}
\put (1.8,17.2) {Blurred image}
\put (17.7,17.2) {DeblurGAN \cite{DeblurGAN}}
\put (36.0,17.2) {FastGyro \cite{mustaniemi2018fast}}
\put (51.0,17.2) {DeepBlind (ours)}
\put (68.0,17.2) {DeepGyro (ours)}
\put (89.0,17.2) {Sharp}
\end{overpic}
\caption{Deblurring results on an image with synthetic linear blur, length 60 pixels. The blur passed to the non-blind methods -- FastGyro and DeepGyro -- is biased,  $\epsilon = [5,3]$ pixels is added to the blur vector in the x and y directions, respectively.}
\vspace{2mm}
%\caption{Deblurring results on an image with synthetic linear blur, length 60 pixels. The blur passed as input to the methods is biased, it is larger than the ground truth by 5 and 3 pixels in the x and  y directions, respectively.}
\label{fig:synthetic}
\end{figure*}

Figure \ref{fig:synthetic} shows the deblurring results on a heavily blurred image. DeepBlind and DeepGyro clearly outperform the rest of the methods. Their performance is comparable to each other, although DeepGyro results to a slightly higher PSNR and SSIM values. The average results for all scenes are summarized in Table \ref{tab:metrics}. DeepGyro benefits from the initial blur estimates, especially when there is significant amount of blur.

\begin{figure*}[!h]
\centering
\begin{overpic}[width=1.00\textwidth]{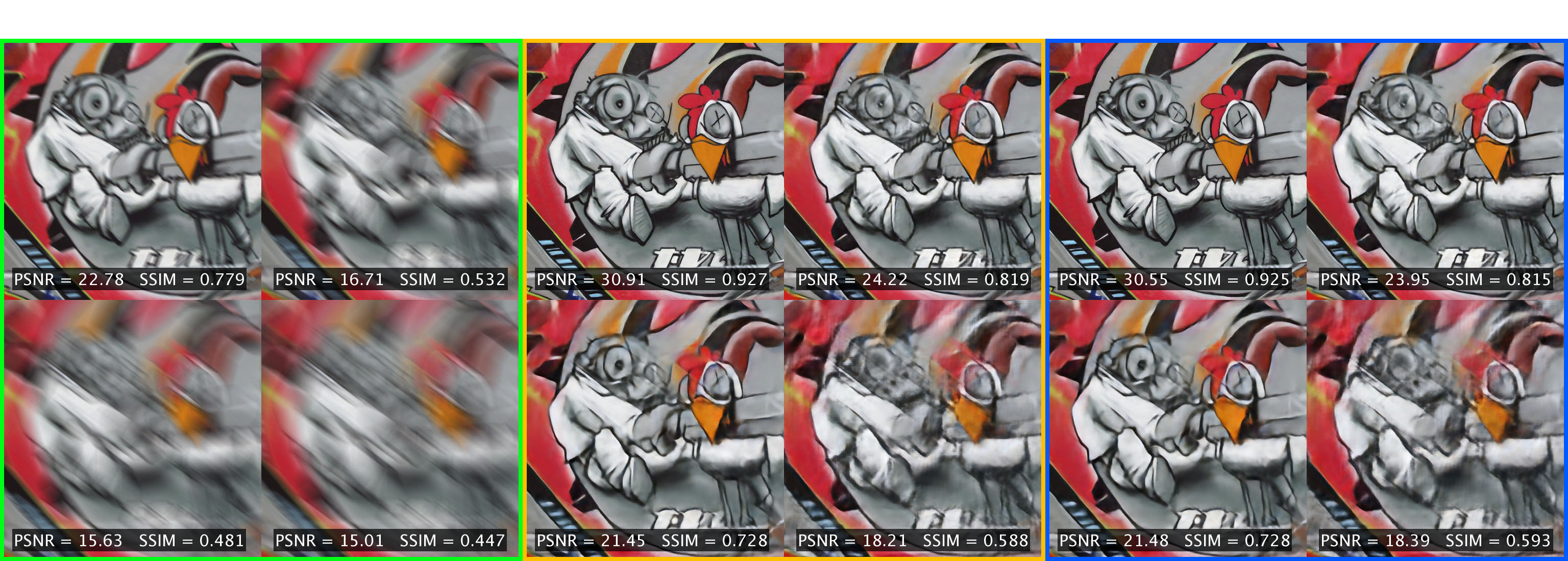}
\put (10.0,34.0) {Blurred images}
\put (44.0,34.0) {DeepGyro (exact)}
\put (75.0,34.0) {DeepGyro (noisy)}
\end{overpic}
\caption{DeepGyro performance for increasing levels of blur. Blurred images (green). Results obtained when passing the exact blur $\epsilon = [0,0]$ as input (orange), and when the blur vector has an error $\epsilon = [5,3]$ pixels (blue). Testing with blur sizes: 10, 40, 60, 80 pixels.}
\vspace{2mm}
\label{fig:blur_size}
\end{figure*}
\begin{figure*}[!h]
\centering
\begin{overpic}[width=1.00\textwidth]{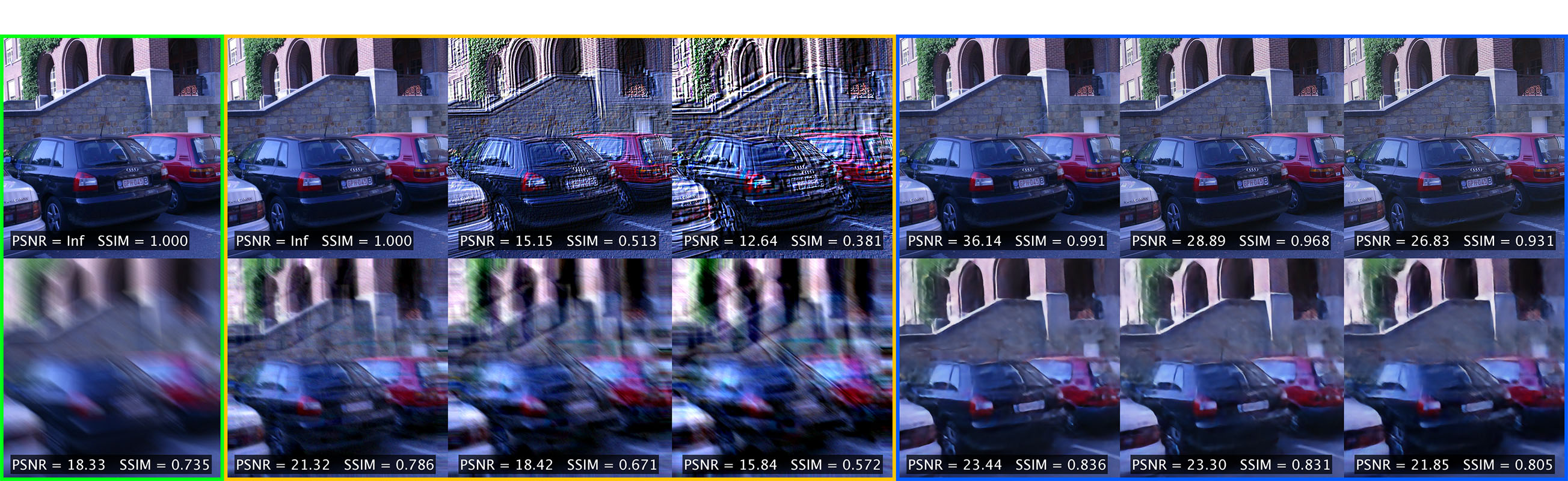}
\put (2.0,29.4) {Input images}
\put (31.0,29.4) {FastGyro \cite{mustaniemi2018fast}}
\put (75.0,29.4) {DeepGyro}
\end{overpic}
\caption{The effects of blur estimation errors on the FastGyro \cite{mustaniemi2018fast} (orange) and DeepGyro (blue). Showing the results when the input is sharp (top row) and when it is blurred (bottom row). The error of the blur is gradually increased $\epsilon = k*[5,3]$ pixels, where $k=0,2,4$.}
\label{fig:blur_error}
\end{figure*}

Figure \ref{fig:blur_size} shows the performance of DeepGyro for increasing levels of motion blur. The method is able to handle extremely strong motion blur. It performs well even when the input blur is not perfect. Figure \ref{fig:blur_error} investigates the effects of blur estimation errors in more detail. Notice that FastGyro \cite{mustaniemi2018fast} is quite sensitive to these errors as there are major ringing artifacts. Another important property of DeepGyro is that it never ruins an already sharp image.

\subsection{Natural blur}
Naturally blurred images were captured with the NVIDIA Shield tablet while simultaneously logging gyroscope at 100 Hz. In this section, we rely on visual assessment since the ground truth sharp images are not available. Figure \ref{fig:natural} shows the deblurring results. The resolution of the images is 512 x 512 pixels. DeepGyro performs consistently better than the other methods. In many cases, DeepBlind leaves some parts of the image blurred. FastGyro \cite{mustaniemi2018fast} is able to recover a lot of details but the artifacts reduce the quality of the image. DeblurGAN \cite{DeblurGAN} struggles with strong motion blur. It also seems to produce a grid-like pattern over the image. We also tested our method on a blurred video sequence. Figure \ref{fig:teaser} (left) shows the result for a single frame with a resolution of 270 x 480 pixels. The deblurring takes around 35 milliseconds on NVIDIA GeForce GTX 1080 GPU. The full video is provided in the supplementary material.

\begin{table*}[!h]
\newcolumntype{.}{D{.}{.}{-1}}
\centering
\setlength{\tabcolsep}{6.3pt}
\caption{Quantitative comparison of deblurring methods on synthetically blurred images: the average PSNR and SSIM metrics for increasing levels of motion blur on the first image from the \textit{graf}, \textit{ubc}, \textit{bikes} and \textit{leuven} image sets \cite{mikolajczyk2005performance}.
DeepGyro* - results when the input blur is exact (blur only caused by rotation).}
\begin{tabular}{lllllllllllll}
\toprule
\multicolumn{1}{c}{Blur size} &
\multicolumn{2}{c}{Blurred image} &
\multicolumn{2}{c}{DeblurGAN \cite{DeblurGAN}} &
\multicolumn{2}{c}{FastGyro \cite{mustaniemi2018fast}} &
\multicolumn{2}{c}{DeepBlind} &
\multicolumn{2}{c}{DeepGyro} &
\multicolumn{2}{c}{DeepGyro*} \\
\cmidrule(r){2-3}
\cmidrule(r){4-5}
\cmidrule(r){6-7}
\cmidrule(r){8-9}
\cmidrule(r){10-11}
\cmidrule(r){12-13}
\multicolumn{1}{l}{(pixels)} & 
\multicolumn{1}{l}{PSNR} &
\multicolumn{1}{l}{SSIM} &
\multicolumn{1}{l}{PSNR} &
\multicolumn{1}{l}{SSIM} &
\multicolumn{1}{l}{PSNR} &
\multicolumn{1}{l}{SSIM} &
\multicolumn{1}{l}{PSNR} &
\multicolumn{1}{l}{SSIM} &
\multicolumn{1}{l}{PSNR} &
\multicolumn{1}{l}{SSIM} &
\multicolumn{1}{l}{PSNR} &
\multicolumn{1}{l}{SSIM}\\ 
\cmidrule(r){2-3}
\cmidrule(r){4-5}
\cmidrule(r){6-7}
\cmidrule(r){8-9}
\cmidrule(r){10-11}
\cmidrule(r){12-13}
10 & 23.41 & 0.792 & 21.46 & 0.694 & 18.55 & 0.567 & 29.10 & 0.920 & 28.91 & 0.918 & \textbf{29.34} & \textbf{0.921} \\
20 & 20.69 & 0.705 & 20.23 & 0.642 & 19.54 & 0.611 & \textbf{27.14} & 0.879 & 26.93 & 0.880 & 27.10 & \textbf{0.880} \\
40 & 18.64 & 0.647 & 18.39 & 0.593 & 19.73 & 0.642 & 24.32 & 0.815 & 24.58 & 0.821 & \textbf{24.74} & \textbf{0.823} \\
60 & 17.61 & 0.617 & 17.27 & 0.565 & 19.18 & 0.633 & 21.92 & 0.746 & 22.55 & 0.757 & \textbf{22.66} & \textbf{0.759} \\
80 & 16.97 & 0.598 & 16.70 & 0.557 & 18.28 & 0.605 & 19.13 & 0.652 & \textbf{20.20} & \textbf{0.684} & 19.97 & 0.681 \\
\bottomrule
\rule{0pt}{1ex}
\end{tabular}
\label{tab:metrics}
\end{table*}

\begin{figure*}
\centering
\begin{overpic}[width=0.87\textwidth]{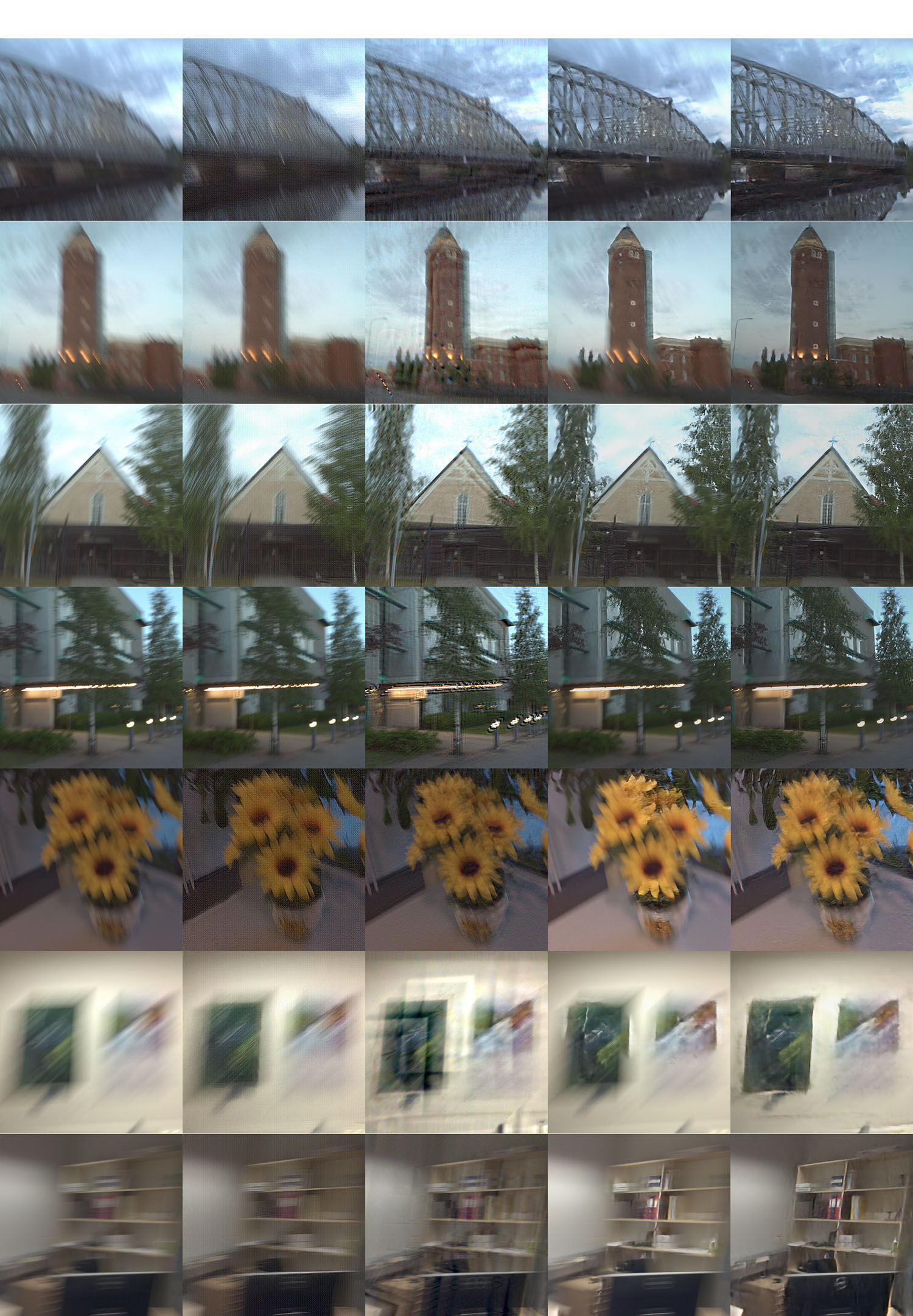}
\put (2.5,97.6) {Blurred image}
\put (15.5,97.6) {DeblurGAN \cite{DeblurGAN}}
\put (30.0,97.6) {FastGyro \cite{mustaniemi2018fast}}
\put (43.0,97.6) {DeepBlind (ours)}
\put (56.5,97.6) {DeepGyro (ours)}
\end{overpic}
\caption{Deblurring images blurred by camera motion. From top to bottom: \textit{bridge, tower, church, entrance, flower, posters, office}.}
\label{fig:natural}
\end{figure*}

\begin{figure*}
\centering
\begin{overpic}[width=1.00\textwidth]{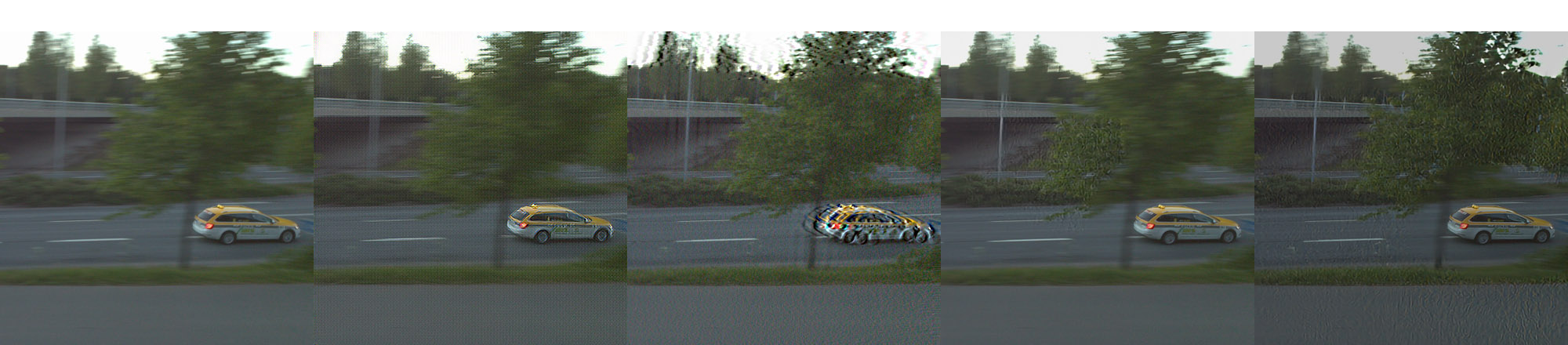}
\put (4,21.0) {Blurred image}
\put (23.0,21.0) {DeblurGAN \cite{DeblurGAN}}
\put (45,21.0) {FastGyro \cite{mustaniemi2018fast}}
\put (62.5,21.0) {DeepBlind (ours)}
\put (82.5,21.0) {DeepGyro (ours)}
\end{overpic}
\caption{Deblurring a dynamic scene.}
\label{fig:dynamic}
\end{figure*}

None of the methods is designed for dynamic scene deblurring. Nevertheless, Figure \ref{fig:dynamic} shows a dynamic scene in which a moving car is tracked by the camera. DeepGyro is able to remove most of the blur caused by the camera motion. The car also remains sharp, although a small area around the car is left blurred. This problem is likely due to the fact that the blur does not vary smoothly across the image (as it would in case of camera motion only).

\begin{figure*}[!h]
\centering
\begin{overpic}[width=1.00\textwidth]{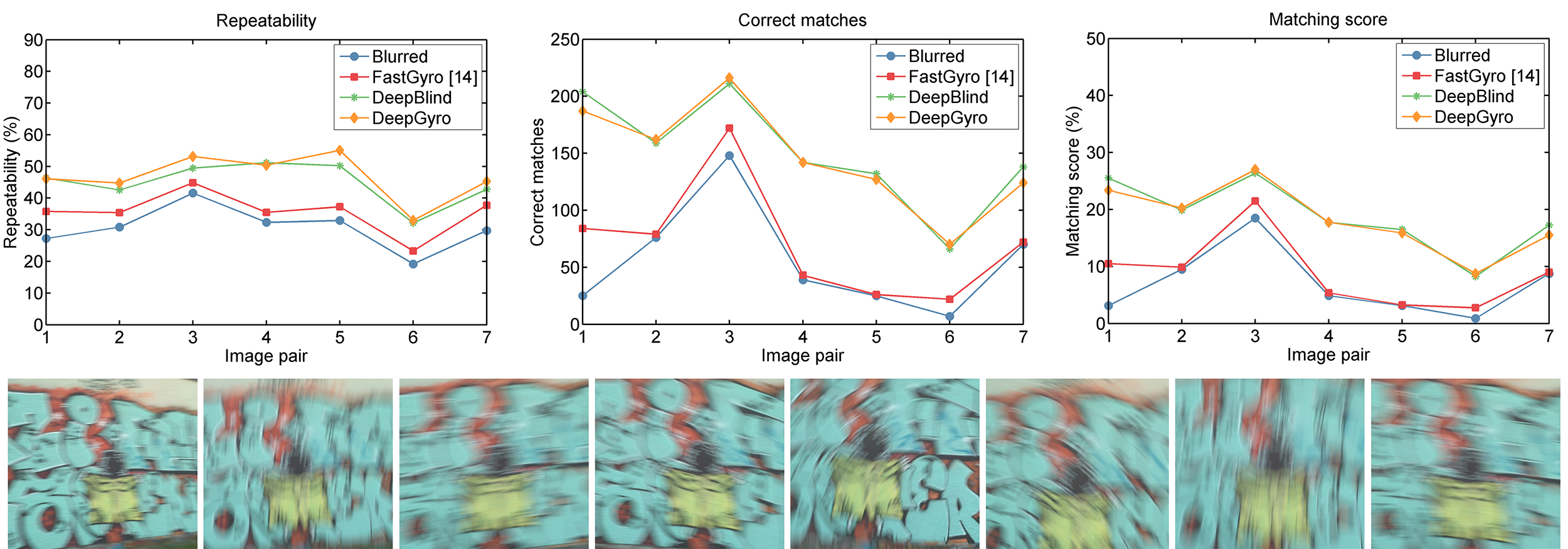}
\end{overpic}
\caption{The evaluation of feature detection and matching. Images used in the experiment (bottom). The left-most image is used as the reference. Repeatability scores computed for each image pair (left).  The overlap criteria is set to 40 \% and the number of detections is fixed to 800. Number of correct nearest neighbour matches (center) and matching scores (right).}
\label{fig:detection_and_matching}
\end{figure*}

The results are generally quite impressive but there is still room for an improvement. The \textit{entrance} scene in Figure \ref{fig:natural} contains bright light sources, which cause some of the pixels to saturate. Consequently, this area is not deblurred. %The problem does not occur when deblurring a cropped version of the same image. 
The light streaks also indicate that the blur is somewhat nonlinear. This will likely reduce the deblurring performance because such images are not present in the training set. The \textit{flower} scene also shows that a significant translation can cause problems when the scene is close. In this case, it is probable that the gyro-based blur field differs too much from the real blur.

\subsection{Feature detection and matching}
Motion blur degrades the performance of existing feature detectors and descriptors \cite{gauglitz2011evaluation}. In this section, we use the proposed methods to improve the robustness against motion blur. Specifically, we use the publicly available implementation of Difference of Gaussian (DoG) detector and SIFT descriptor \cite{vedaldi2007open}. The experiment is performed on real-world images with spatially-variant motion blur. The images are shown in Figure \ref{fig:detection_and_matching}.

For the evaluation, we need to know the ground truth homography between the images. It defines the mapping of image points in the first and second image given a planar scene. Normally, the homography can be estimated by selecting corresponding points from the images. In this case, the images are blurred, which makes it difficult to select the points accurately. To solve the issue, we adapt the procedure from \cite{mustaniemi2018fast}. The idea is to capture a burst of images while alternating short and long exposure time. The corresponding points are easier to select from the short exposure images, which are sharp but noisy. The blurred images in Figure \ref{fig:detection_and_matching} also suffer from the rolling shutter distortion. Therefore, a homography cannot necessarily perfectly define the mapping of image points. Nevertheless, we concluded that the homographies are sufficiently accurate for this experiment.

To evaluate feature detection, we compute the repeatability, i.e. how well does the detector identify the corresponding image regions. It is well known that the repeatability criteria might favor detectors that return many keypoints. To eliminate this issue, we fix the number of detections. The results of the experiment are shown in Figure \ref{fig:detection_and_matching}. DeepGyro and DeepBlind clearly outperform the standard detector without deblurring as well as the FastGyro \cite{mustaniemi2018fast}.

For the feature matching evaluation, we compute the number of correct matches and the matching score. The nearest neighbour in the descriptors space corresponds to a match. The matching score is the ratio between the number of correct matches and the smaller number of detected features in the pair of images. The results of the experiment are shown in Figure \ref{fig:detection_and_matching}. Again, the performance of DeepGyro and DeepBlind is superior compared to the other approaches. 

In this experiment, the performance of DeepGyro and DeepBlind is close to equal. The scene in Figure \ref{fig:detection_and_matching} has a lot of texture, which helps especially the DeepBlind. The information from the gyroscope seems to be redundant when DeepBlind performs well.

%-------------------------------------------------------------------------
\section{Conclusion}
We proposed a deblurring method that is first to pass gyroscope readings to a CNN.
The network learns that gyro-based blur estimates are noisy, which allows it to avoid deblurring artifacts common to non-blind deconvolution methods.
The evaluation shows that the method handles extreme and spatially-variant motion blur in real-time, unlike existing methods, and that
it does not damage images that are sharp.
Many of the aforementioned benefits are made possible by the proposed data generation scheme, which utilizes gyroscope readings to produce realistic training data. 
Finally, it was demonstrated that the method improves performance of existing feature detectors and descriptors against the motion blur.

%------------------------------------------------------------------------
\section*{Acknowledgment}
The work has been financially supported by the FiDiPro programme of Business Finland and J. Matas was supported by OP VVV MEYS project CZ.02.1.01/ 0.0/0.0/16\_019/0000765 ''Research Center for Informatics''.

%\clearpage

{\small
\bibliographystyle{ieee}
\bibliography{}
%\bibliography{bibfile}
}

\end{document}